%% file: 3D_car_arxiv.tex
\documentclass[10pt,twocolumn,letterpaper]{article}

\usepackage{cvpr}
\usepackage{times}
\usepackage{epsfig}
\usepackage{graphicx}
\usepackage{amsmath}
\usepackage{amssymb}

\usepackage{helvet}
\usepackage{courier}
\usepackage{bm}
\usepackage{color}

\usepackage{bbm}
\usepackage{epstopdf}
\usepackage{caption}
\usepackage{subcaption}
\usepackage{enumitem}
\usepackage{calc}
\usepackage{multirow}
\usepackage{xspace}
\usepackage{booktabs}
\usepackage{mathrsfs}
\usepackage{array}
\usepackage[font=small,skip=4pt]{caption}
\usepackage{graphicx}
\usepackage{threeparttable}
\usepackage{dsfont}
\usepackage{authblk}

\usepackage{verbatim}

\newcommand{\figref}[1]{Fig\onedot~\ref{#1}}
\newcommand{\equref}[1]{Eq\onedot~\eqref{#1}}
\newcommand{\secref}[1]{Sec\onedot~\ref{#1}}
\newcommand{\tabref}[1]{Tab\onedot~\ref{#1}}

\newcommand{\ve}[1]{{\mathbf #1}} 
\newcommand{\hua}[1]{{\mathcal #1}}

\newcommand{\by}[2]{\ensuremath{#1 \! \times \! #2}}
\newcommand{\thickhline}{%
    \noalign {\ifnum 0=`}\fi \hrule height 1pt
    \futurelet \reserved@a \@xhline
}

\DeclareRobustCommand\onedot{\futurelet\@let@token\@onedot}
\def\onedot{\ifx\@let@token.\else.\null\fi\xspace}
\def\eg{\emph{e.g.}}

\def\ie{\emph{i.e.}}

\def\etc{\emph{etc}\onedot}

\def\wrt{w.r.t\onedot}

\usepackage[breaklinks=true,bookmarks=false]{hyperref}

\cvprfinalcopy 

\def\cvprPaperID{****} 

\begin{document}

\title{ApolloCar3D: A Large 3D Car Instance Understanding Benchmark for Autonomous Driving}




\author{Xibin Song$^{1,2}$, Peng Wang$^{1,2}$, Dingfu Zhou$^{1,2}$, Rui Zhu$^3$, Chenye Guan$^{1,2}$, \\ Yuchao Dai$^{4}$, Hao Su$^3$, Hongdong Li$^{5,6}$ and Ruigang Yang$^{1,2}$}

\affil{$^1$Baidu Research \quad $^2$National Engineering Laboratory of Deep Learning Technology and Application, China \quad $^3$University of California, San Diego \quad
$^4$ Northwestern Polytechnical University, Xi'an, China\quad
$^5$ Australian National University, Australia \quad $^6$Australian Centre for Robotic Vision, Australia\\
\tt\small \{songxibin,wangpeng54,zhoudingfu,guanchenye,yangruigang\}@baidu.com,\\
\tt\small \{rzhu,haosu\}@eng.ucsd.edu, daiyuchao@gmail.com and hongdong.li@anu.edu.au}

\maketitle

\begin{abstract}
Autonomous driving has attracted remarkable attention from both industry and academia. An important task is to estimate 3D properties (\eg\ translation, rotation and shape) of a moving or parked vehicle on the road.  This task, while critical, is still under-researched in the computer vision community -- partially owing to the lack of large scale and fully-annotated 3D car database suitable for autonomous driving research. In this paper, we contribute the first large-scale database suitable for 3D car instance understanding -- \texttt{ApolloCar3D}. The dataset contains 5,277 driving images and over 60K car instances, where each car is fitted with an industry-grade 3D CAD model with absolute model size and semantically labelled keypoints. This dataset is above 20$\times$ larger than PASCAL3D+~\cite{xiang2014beyond} and KITTI~\cite{geiger2012we}, the current state-of-the-art. To enable efficient labelling in 3D, we build a pipeline by considering 2D-3D keypoint correspondences for a single instance and 3D relationship among multiple instances. Equipped with such dataset, we build various baseline algorithms with the state-of-the-art deep convolutional neural networks. Specifically, we first segment each car with a pre-trained Mask R-CNN~\cite{he2017mask}, and then regress towards its 3D pose and shape based on a deformable 3D car model with or without using semantic keypoints. We show that using keypoints significantly improves fitting performance. Finally, we develop a new 3D metric jointly considering 3D pose and 3D shape, allowing for comprehensive evaluation and ablation study. 
By comparing with human performance we suggest several future directions for further improvements.
\end{abstract}

\input{intro.tex}
\input{data_arxiv.tex}
\input{approach.tex}
\input{kp_based.tex}
\input{exp_arxiv.tex}

\vspace{-0.1in}
\section{Conclusion}
This paper presents by far the largest and growing dataset (namely \texttt{ApolloCar3D}) for instance-level 3D car understanding in the context of autonomous driving. It is built upon industrial-grade high-precision 3D car models fitted to car instances captured in real world scenarios. Complementing existing related datasets \eg\ ~\cite{geiger2012we}, we hope this new dataset could  serve as a long-standing benchmark facilitating future research on 3D pose and shape recovery.

In order to efficiently annotate complete 3D object properties, we have developed a context-aware 3D annotation pipeline, as well as two baseline algorithms for evaluation.  We have also conducted carefully designed human performance study, which reveals that there is still a visible gap between machine performance and that of human's, motivating and suggesting promising future directions.  More importantly, built upon the publicly available ApolloScape dataset~\cite{huang2018apolloscape}, our \texttt{ApolloCar3D} dataset contains multitude of data sources including stereo, camera pose, semantic instance label, per-pixel depth ground truth, and moving videos. Working with our data enables training and evaluation of a wide range of other vision tasks, \eg\ stereo vision, model-free depth estimation, and optical flow \etc, under real scenes. 

\section{Acknowledgement}

The authors gratefully acknowledge He Jiang from Baidu Research for car visualization using obtained poses. Meanwhile, the authors also gratefully acknowledge Maximilian Jaritz from University of California, San Diego for counting car numbers of the proposed dataset.

{\small
\bibliographystyle{ieee}
\bibliography{CarFitting}
}

\begin{appendix}
\section{Keypoints Definition}
Here we show the definitions of 66 semantic keypoints (Fig.~\ref{fig:kp_definition}).
\begin{itemize}
    \item 0: Top left corner of left front car light;
    \item 1: Bottom left corner of left front car light;
    \item 2: Top right corner of left front car light;
    \item 3: Bottom right corner of left front car light;
    \item 4: Top right corner of left front fog light;
    \item 5: Bottom right corner of left front fog light;
    \item 6: Front section of left front wheel;
    \item 7: Center of left front wheel;
    \item 8: Top right corner of front glass;
    \item 9: Top left corner of left front door;
    \item 10: Bottom left corner of left front door;
    \item 11: Top right corner of left front door;
    \item 12: Middle corner of left front door;
    \item 13: Front corner of car handle of left front door;
    \item 14: Rear corner of car handle of left front door;
    \item 15: Bottom right corner of left front door;
    \item 16: Top right corner of left rear door;
    \item 17: Front corner of car handle of left rear door;
    \item 18: Rear corner of car handle of left rear door;
    \item 19: Bottom right corner of left rear door;
    \item 20: Center of left rear wheel;
    \item 21: Rear section of left rear wheel;
    \item 22: Top left corner of left rear car light;
    \item 23: Bottom left corner of left rear car light;
    \item 24: Top left corner of rear glass;
    \item 25: Top right corner of left rear car light;
    \item 26: Bottom right corner of left rear car light;
    \item 27: Bottom left corner of trunk;
    \item 28: Left corner of rear bumper;
    \item 29: Right corner of rear bumper;
    \item 30: Bottom right corner of trunk;
    \item 31: Bottom left corner of right rear car light;
    \item 32: Top left corner of right rear car light;
    \item 33: Top right corner of rear glass;
    \item 34: Bottom right corner of right rear car light;
    \item 35: Top right corner of right rear car light;
    \item 36: Rear section of right rear wheel;
    \item 37: Center of right rear wheel;
    \item 38: Bottom left corner of right rear car door;
    \item 39: Rear corner of car handle of right rear car door;
    \item 40: Front corner of car handle of right rear car door; \item 41: Top left corner of right rear car door;
    \item 42: Bottom left corner of right front car door;
    \item 43: Rear corner of car handle of right front car door; \item 44: Front corner of car handle of right front car door; \item 45: Middle corner of right front car door;
    \item 46: Top left corner of right front car door;
    \item 47: Bottom right corner of right front car door;
    \item 48: Top right corner of right front car door;
    \item 49: Top left corner of front glass;
    \item 50: Center of right front wheel;
    \item 51: Front section of right front wheel;
    \item 52: Bottom left corner of right fog light;
    \item 53: Top left corner of right fog light;
    \item 54: Bottom left corner of right front car light;
    \item 55: Top left corner of right front car light;
    \item 56: Bottom right corner of right front car light;
    \item 57: Top left corner of right front car light;
    \item 58: Top right corner of front license plate;
    \item 59: Top left corner of front license plate;
    \item 60: Bottom left corner of front license plate;
    \item 61: Bottom right corner of front license plate;
    \item 62: Top left corner of rear license plate;
    \item 63: Top right corner of rear license plate;
    \item 64: Bottom right corner of rear license plate;
    \item 65: Bottom left corner of rear license plate.
\end{itemize}
\end{appendix}

\end{document}

%% file: intro.tex
\section{Introduction}
\label{sec:intro}

\begin{figure}[t]

  \centering
  \includegraphics[width=0.43\textwidth]{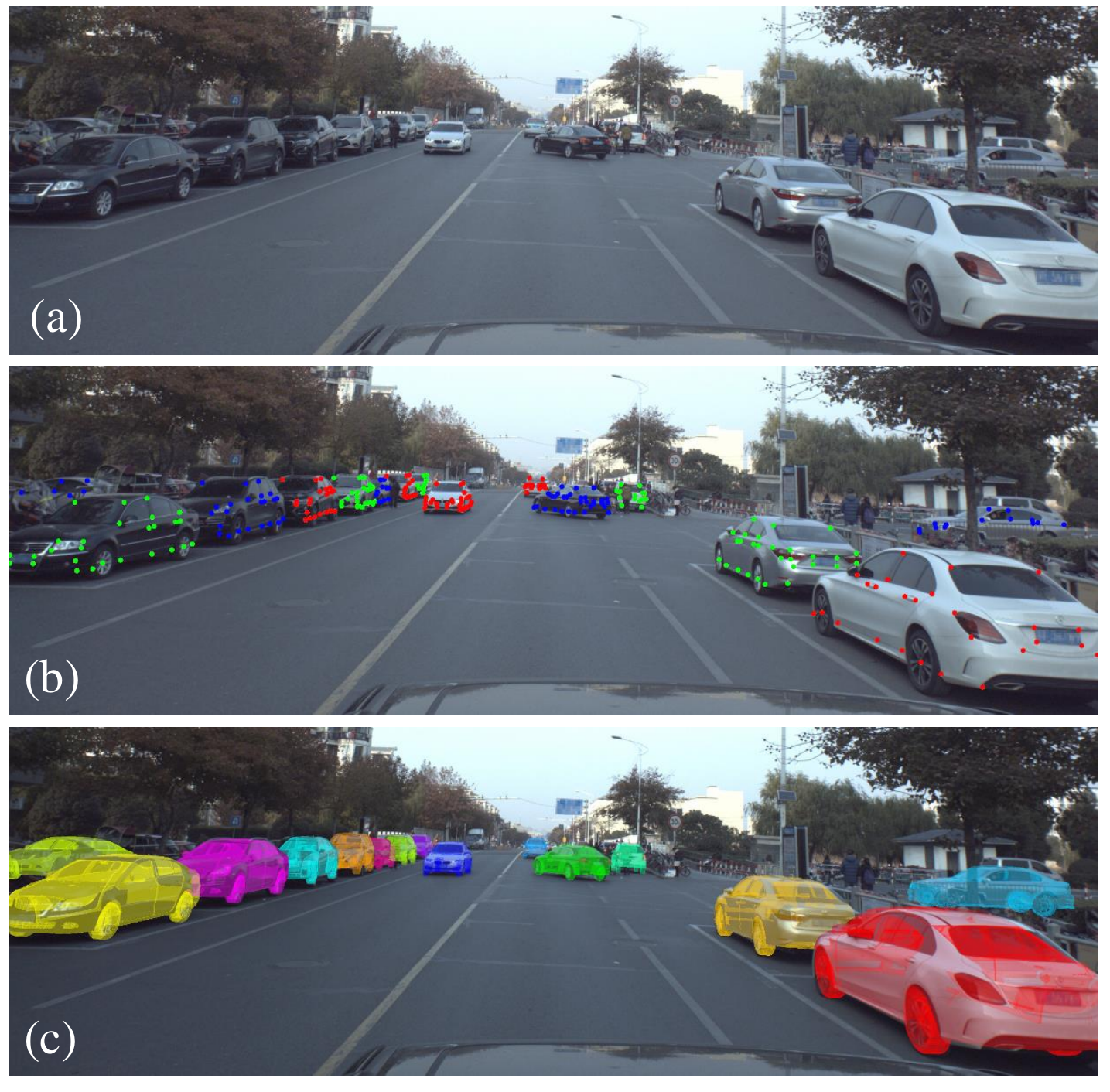}

 \caption{An example of our dataset, where (a) is the input color image, (b) illustrates the labeled 2D keypoints, (c) shows the 3D model fitting result with labeled 2D keypoints.}
\label{fig:tittle_img}
\end{figure}
Understanding 3D properties of objects from an image, \ie\ to recover objects' 3D pose and shape, is an important task of computer vision, as illustrated in \figref{fig:tittle_img}. This task is also called ``inverse-graphics''~\cite{kulkarni2015deep}, solving which would enable a wide range of applications in vision and robotics, such as robot navigation~\cite{leonard2012directed}, visual recognition~\cite{engelmann2016joint}, and human-robot interaction~\cite{canal2016real}.
Among them, autonomous driving (AD) is a prominent topic which holds great potential in practical applications. Yet, in the context of AD the current leading technologies for 3D object understanding mostly rely on high-resolution LiDAR sensor ~\cite{liang2018deep}, rather than regular camera or image sensors. 

\begin{table*}[t]
\center
\renewcommand*{\arraystretch}{1.2}
\fontsize{8.5}{9}\selectfont
\setlength\tabcolsep{1.5pt}
\begin{tabular}{l|c|c|c|c|c|c|c|c}
\toprule[0.13em]
\multicolumn{1}{l|}{Dataset}                     & Image source & 3D property    & Car keypoints ($\#$)  & Image ($\#$) &  Average cars/image  &  Maximum cars/image & Car models $\#$  & Stereo \\
\hline
\multicolumn{1}{l|}{3DObject~\cite{savarese20073d}}         & Control      &  complete 3D    & No & 350   & 1  & 1  &  10   & No \\
\multicolumn{1}{l|}{EPFL Car~\cite{ozuysal2009pose}}        & Control      &  complete 3D    & No & 2000  & 1  & 1  &  20   & No \\
\multicolumn{1}{l|}{PASCAL3D+~\cite{xiang2014beyond}}        & Natural      & complete 3D     & No & 6704  &  1.19  &  14   &  10    & No \\
\multicolumn{1}{l|}{ObjectNet3D~\cite{xiang2016objectnet3d}}& Natural      & complete 3D     & Yes (14) & 7345 & 1.75 & 2  & 10 & No \\
\multicolumn{1}{l|}{KITTI~\cite{geiger2012we}}              & Self-driving &3D bbox $\&$ ori. & No & 7481 & 4.8  & 14   & 16     & Yes \\
\hline
\multicolumn{1}{l|}{ApolloCar3D}                            & Self-driving & industrial 3D     & Yes (66) & 5277  & 11.7 & 37 & 79 & Yes \\
\toprule[0.13 em]
\end{tabular}

\caption{Comparison between our dataset and existing datasets with 3D car labels. ``complete 3D'' means fitting with 3D car model.}
\label{Tab:other3Ddataset}
\end{table*}

However, we argue that there are multitude drawbacks in using LiDAR, hindering its further up-taking. The most severe one is that the recorded 3D LiDAR points are at best a sparse coverage of the scene from front view~\cite{geiger2012we}, especially for distant and absorbing regions. Since it is crucial for a self-driving car to maintain a safe breaking distance, 3D understanding from a regular camera remains a promising and viable approach attracting significant amount of research from the vision community ~\cite{chen2016monocular,tran2015learning}.  

The recent tremendous success of deep convolutional network~\cite{he2017mask} in solving various computer vision tasks is built upon the availability of massive carefully annotated training datasets, such as ImageNet~\cite{deng2009imagenet} and MSCOCO~\cite{lin2014microsoft}. Acquiring large-scale training datasets however is an extremely laborious and expensive endeavour, and the community is especially lacking of fully annotated datasets of 3D nature. For example, for the task of 3D car understanding for autonomous driving, the availability of datasets is severely limited. Take KITTI~\cite{geiger2012we} for instance. Despite being the most popular dataset for self-driving, it has only about $200$ labelled 3D cars yet in the form of bounding box only, without detailed 3D shape information flow~\cite{menze2015object}. Deep learning methods are generally hungry for massive labelled training data, yet the sizes of currently available 3D car datasets are far from adequate to capture various appearance variations, \eg\ occlusion, truncation, and lighting. For other datasets such as PASCAL3D+~\cite{xiang2014beyond} and ObjectNet3D~\cite{xiang2016objectnet3d}, while they contain more images, the car instances therein are mostly isolated, imaged in a controlled lab setting thus are unsuitable for autonomous driving.  

To rectify this situation, we propose a large-scale 3D instance car dataset built from real images and videos captured in complex real-world driving scenes in multiple cities. Our new dataset, called \texttt{ApolloCar3D}, is built upon the publicly available ApolloScape dataset~\cite{huang2018apolloscape} and targets at 3D car understanding research in self-driving scenarios. Specifically, we select $5,277$ images from around 200K released images in the semantic segmentation task of ApolloScape, following several principles such as (1) containing sufficient amount of cars driving on the street, (2) exhibiting large appearance variations, (3) covering multiple driving cases at highway, local, and intersections. In addition, for each image, we provide a stereo pair for obtaining stereo disparity; and for each car, we provide 3D keypoints such as corner of doors and headlights, as well as realistic 3D CAD models with an absolute scale. An example is shown in~\figref{fig:tittle_img}(b). We will provide details about how we define those keypoints and label the dataset in \secref{sec:data}.

Equipped with \texttt{ApolloCar3D}, we are able to directly apply supervised learning to train a 3D car understanding system from images, instead of making unnecessary compromises falling back to weak-supervision or semi-supervision like most previous works do, \eg\ 3D-RCNN~\cite{kundu20183d} or single object 3D recovery~\cite{wang20183d}.  

To facilitate future research based on our \texttt{ApolloCar3D} dataset, we also develop two 3D car understanding algorithms, to be used as new baselines in order to benchmark future  contributed algorithms. Details of our baseline algorithms will be described in following sections.  

Another important contribution of this paper is that we propose a new evaluation metric for this task, in order to to jointly measure the quality of both 3D pose estimation and shape recovery. We referred to our new metric as ``Average 3D precision (A3DP)", as it is inspired by the AVP metric (average viewpoint precision) for PASCAL3D+~\cite{xiang2014beyond} which however only considers 3D pose. In addition, we supply multiple true positive thresholds similar to MS COCO~\cite{lin2014microsoft}.

The contributions of this paper are summarized as:
{ \begin{itemize}
    \item A large-scale and growing 3D car understanding dataset for autonomous driving, \ie~\texttt{ApolloCar3D}, which complements existing public 3D object datasets. 
    \item A novel evaluation metric, \ie\ A3DP, which jointly considers both 3D shape and 3D pose thus is more appropriate for the task of 3D instance understanding.
    \item Two new baseline algorithms for 3D car understanding, which outperform several state-of-the-art 3D object recovery methods.
    \item Human performance study, which points out promising future research directions.\end{itemize}
}

%% file: data_arxiv.tex
\section{ApolloCar3D Dataset}
\label{sec:data}

\begin{figure*}[t]
 \includegraphics[width=0.99\linewidth]{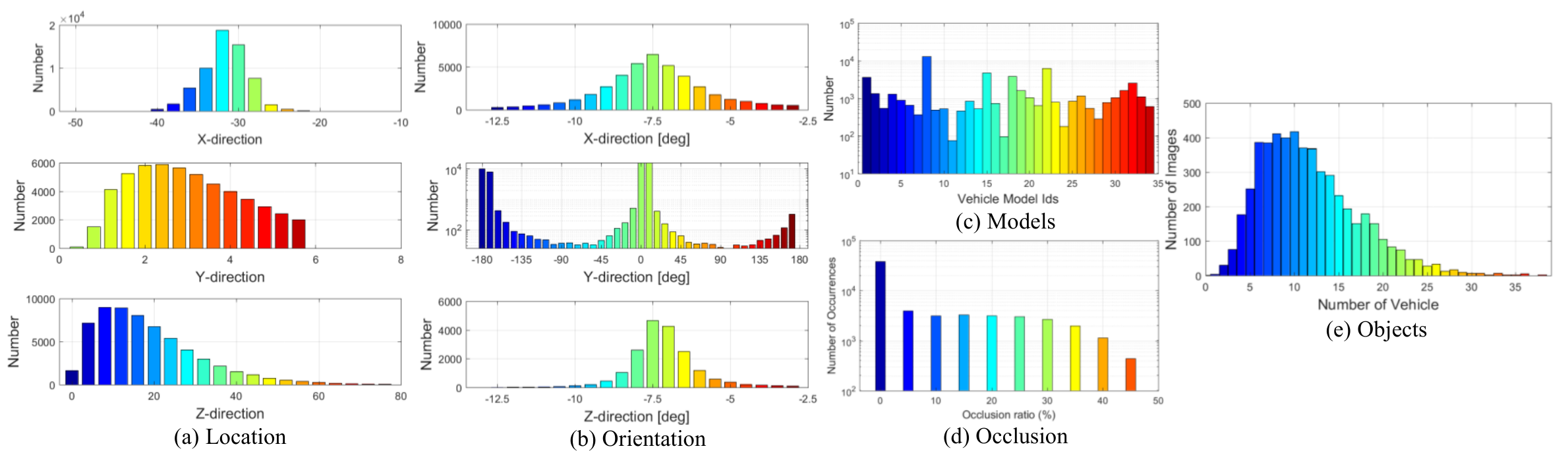}
\caption{Car occurrence and object geometry statistics in \texttt{ApolloCar3D}. (a) and (b) illustrate the translation and orientation distribution of all the vehicles. (c) - (e) describe the distribution of vehicle type, occlusion ratio, and number of vehicles per image. Specifically, the Y-axis in all the figures represents the occurrences of vehicles.}
\label{fig:stat}
\end{figure*}
\paragraph{Existing datasets with 3D object instances.}
Previous datasets for 3D object understanding are often very limited in scale, or with partial 3D properties only, or contains few objects per image ~\cite{leibe2003analyzing,thomas2006towards,savarese20073d,moreels2007evaluation,ozuysal2009pose,lopez2010icaro}. For instance, 3DObject~\cite{savarese20073d} has only 10 instances of cars. The EPFL Car~\cite{ozuysal2009pose} has 20 cars under different viewpoints but was captured in a controlled turntable rather than in real scenes.

To handle more realistic cases from non-controlled scenes, datasets~\cite{lim2013parsing} with natural images collected from Flickr~\cite{mcauley2012image} or indoor scenes~\cite{dai2017scannet} with Kinect are extended to 3D objects~\cite{russell2009building}. The IKEA dataset~\cite{lim2013parsing} labelled a few hundreds indoor images with 3D furniture models. PASCAL3D+~\cite{xiang2014beyond} labelled the 12 rigid categories in PASCAL VOC 2012~\cite{pascal-voc-2012} images with CAD models. ObjectNet3D~\cite{xiang2016objectnet3d} proposed a much larger 3D object dataset with images from ImageNet~\cite{deng2009imagenet} with 100 categories. These datasets, while useful, are not designed for autonomous driving scenarios. 
To the best of our knowledge, the only real-world dataset that partially meets our requirement is the KITTI dataset~\cite{geiger2012we}. Nonetheless, KITTI only labels each car by a rectangular bounding box, and lacks fine-grained semantic keypoint labels (\eg\ window, headlight). One exception is the work of ~\cite{Menze2015ISA}, yet it falls short in the number of 200 labelled images, and their car parameters are not publicly available.

In this paper, as illustrated in \figref{fig:tittle_img}, we offer to the community the first large-scale and  fully 3D shape labelled dataset with 60K+ car instances, from 5,277 real-world images, based on 34 industry-grade 3D CAD car models.  Moreover, we also provide the corresponding stereo image pairs and accurate 2D keypoint annotations. \tabref{Tab:other3Ddataset} gives a comparison of key properties of our dataset versus existing ones for 3D object instance understanding.

\subsection{Data Acquisition}
We acquire images from the ApolloScape dataset~\cite{huang2018apolloscape} due to its high resolution (3384 $\times$ 2710), large scale ($\geq$140K semantically labelled images), and complex driving conditions. From the dataset, we carefully select images satisfying our requirements as stated in \secref{sec:intro}. Specifically, we select images from their labelled videos of 4 different cites satisfying (1) relatively complex environment, (2) interval between selected images $\geq$ 10 frames. After picking images from the whole dataset using their semantic labels, in order to have more diversity, we prune all images manually, and further select ones which contain better variation of car scales, shapes, orientations, and mutual occlusion between instances, yielding 5,277 images for us to label.

For 3D car models, we look for highly accurate shape models, \ie\ the offset between the boundary of re-projected model and manually labelled mask is less than $3px$ on average. However, 3D car meshes in ShapeNet~\cite{chang2015shapenet} are still not accurate enough for us, and it is too costly to fit each 3D model in the presence of heavy occlusion, as shown in \figref{fig:tittle_img}. Therefore, to ensure the quality (accuracy) of 3D models, we hired online model makers to manually build corresponding 3D models given parameters of absolute shape and scale of certain car type. Overall, we build 34 real models including sedan, coupe, minivan, SUV, and MPV, which has covered the majority of car models and types in the market.

\subsection{Data Statistics}
In \figref{fig:stat}, we provide statistics for the labelled cars \wrt translation, orientation, occlusion, and model shape. Compared with KITTI~\cite{geiger2012we}, \texttt{ApolloCar3D} contains significantly larger amount of cars that are at long distance, under heavy occlusions, and these cars are distributed diversely in space. From \figref{fig:stat}(b), the orientation follows a similar distribution, where the majority of cars on road are driving towards or backwards the data acquisition car. In \figref{fig:stat}(c), we show distribution \wrt car types, where sedans have the most frequent occurrences. The object distribution per image in \figref{fig:stat}(e) shows that most of the images contain more than 10 labeled objects.

\section{Context-aware 3D Keypoint Annotation}
\label{sec:keypoints_anno}
\begin{figure}
    \centering
    \includegraphics[width=\linewidth]{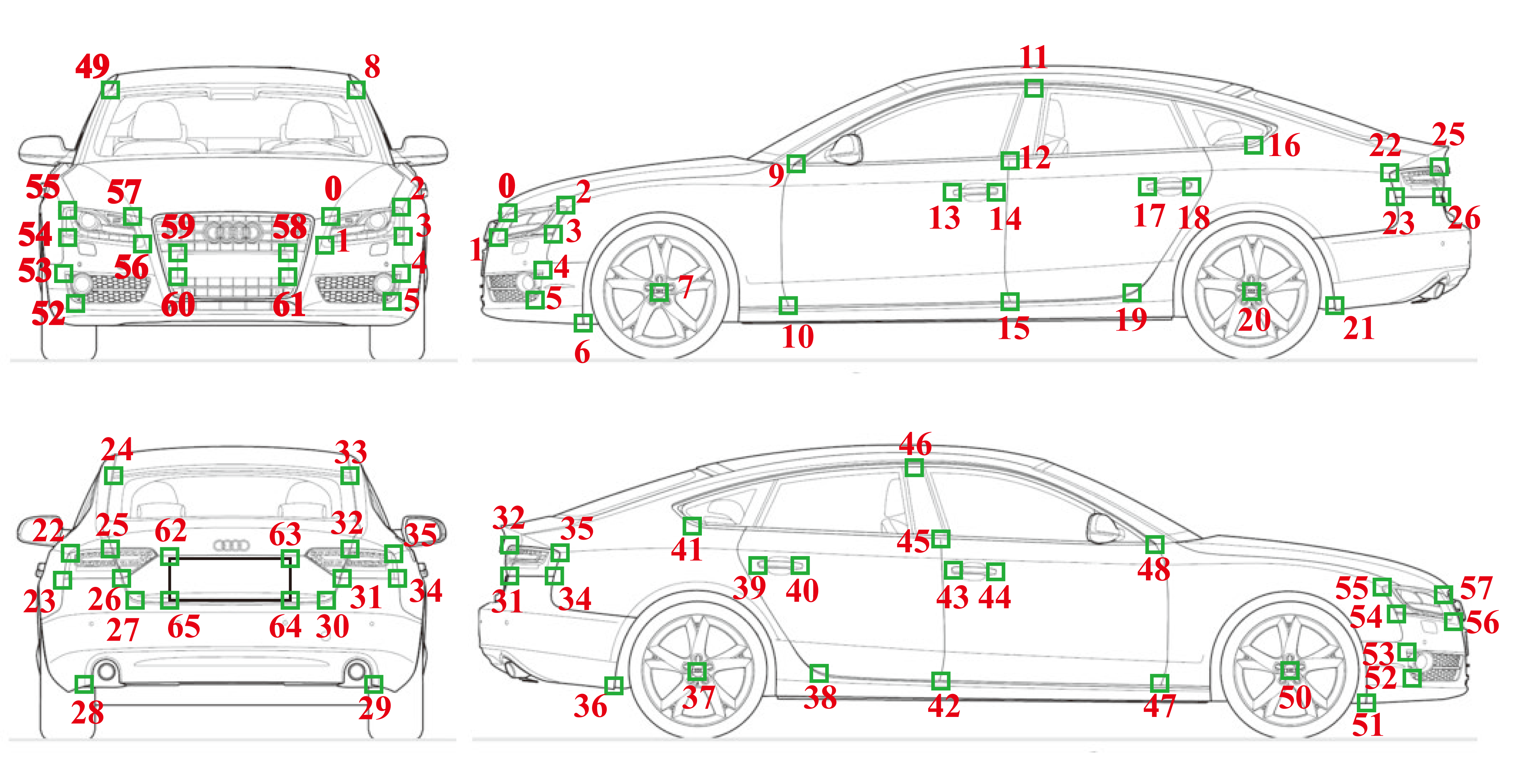}
    \caption{3D keypoints definition for car models. 66 keypoints are defined for each model.}
    \label{fig:kp_definition}
\end{figure}
\begin{figure*}
    \centering
    \includegraphics[width=0.9\linewidth]{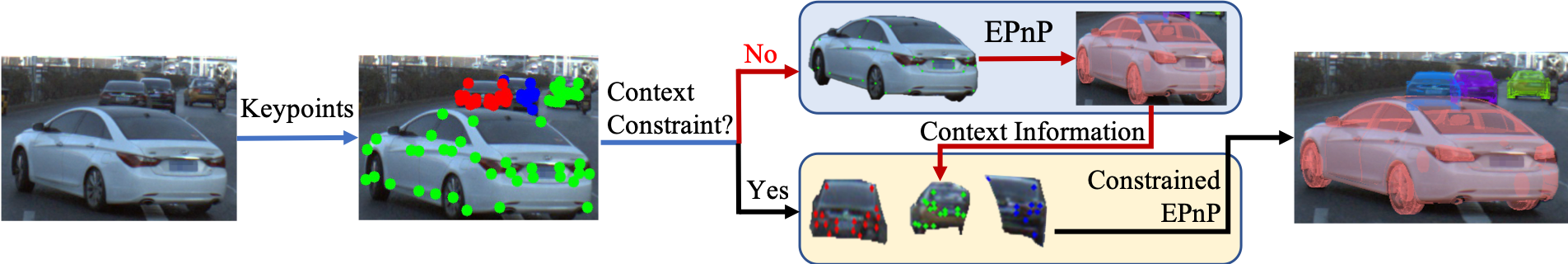}
    \caption{The pipeline for ground truth pose label generation based on annotated 2D and 3D keypoints.}
    \label{fig:label_pipeline}
\end{figure*}




Thanks to the high quality 3D models that we created, we develop an efficient machine-aided semi-automatic keypoint annotation process. Specifically, we only ask human annotators to click on a set of pre-defined keypoints on the object of interest in each image.  Afterwards, the EPnP algorithm~\cite{lepetit2009epnp} is employed to automatically recover the pose and model of the 3D car instance by minimizing re-projection error. RANSAC ~\cite{fischler1981random} is used handle outliers or wrong annotations. While only a handful of keypoints can be sufficient solve the EPnP problem, we define $66$ semantic keypoints in our dataset, as shown in \figref{fig:kp_definition}, which has much higher density than most previous car datasets ~\cite{tulsiani2015viewpoints,miao2016robust}. The redundancy enables more accurate and robust shape-and-pose registration. We will show the definition of each semantic keypoint in appendix.

\noindent\textbf{Context-aware annotation.} In the presence of severe occlusions, for which RANSAC also fails, we develop a context-aware annotation process by enforcing co-planar constraints between one car and its neighboring cars. By doing this, we are able to propagate information among neighboring cars, so that we jointly solve for their poses with context-aware constraints.


Formally, the objective for a single car pose estimation is
\begin{align}
   \hua{E}_{PnP}(\ve{p}, \hua{S}) = \sum_{[\ve{x}^3_k, %
   k]\in \hua{S}} \ve{v}_k\|\pi(\ve{K}, \ve{p},\ve{x}_k^3) - \ve{x}_k \|_2,
    \label{eqn:pnp}
\end{align}
where $\ve{p}\in \ve{S}\ve{E}(3), \hua{S}\in \{S_1, \cdots, S_m\}$ indicate the pose and shape of a car instance respectively. Here, $m$ is the number of models. $\ve{v}$ is a vector indicating whether the $k_{th}$ keypoint of the car has been labelled or not. $\ve{x}_k$ is the labelled 2D keypoint coordinate on the image. $\pi(\ve{p},\ve{x}^3_k)$ is a perspective projection function projecting the correspondent 3D keypoint $\ve{x}^3_k$ on the car model given $\ve{p}$ and camera intrinsic $\ve{K}$.

Our context-aware co-planarity constraint is formulated as:
\begin{align}
    \hua{E}_{N}(\ve{p}, \hua{S}, \ve{p}_n, \hua{S}_n) &=  [(\alpha_{\ve{p}} -  \alpha_{\ve{p}_n})^2  +
     (\beta_{\ve{p}} -  \beta_{\ve{p}_n})^2  \nonumber\\
     &+ ((y_{\ve{p}} - h_{\hua{S}}) - (y_{\ve{p}_n} - h_{\hua{S}_n}))^2],
\end{align}
where $n$ is a spatial neighbor car,  $\alpha_{\ve{p}}$ is roll component of $\ve{p}$, and $h_{\hua{S}}$ is the height of the car given its shape $\hua{S}$.

The total energy to be minimized for finding car pose and shape in image $I$ is defined as:
\begin{align}
    \hua{E}_{I} &= \sum_{c=1}^{C} \{\hua{E}_{PnP}(\ve{p}_c,\hua{S}_c) + \nonumber\\ &B(\hua{K}_c)\sum_{n \in \hua{N}_c}\hua{E}_{N}(\ve{p}_c, \hua{S}_c, \ve{p}_n, \hua{S}_n )\},
\label{eqn:pose_energy}
\end{align}
where $c$ is the index of cars in the image, $B(\hua{K}_c)$ is a binary function indicating whether car $c$ needs to borrow pose information from neighbor cars, and $\hua{K} = \{\ve{x}^2_k\}$ is the set of labelled 2D keypoints of the car. $\hua{N}_c = N(c, \ve{M}, \kappa)$ is the set of rich annotated neighboring cars of $c$ using instance mask $\ve{M}$, and $\kappa$ is the maximum number of neighbors we use.


To judge whether a car needs to use contextual constrains, we define the condition $B(\hua{K}_c)$ in Eq.~(3) for a car instance as the number of annotated keypoints is greater than 6, and the labelled keypoints are lying on more than two predefined car surfaces (detailed in tab.~\ref{tab:correspondence}).

\begin{table}[t]
\newcommand{\tabincell}[2]{\begin{tabular}{@{}#1@{}}#2\end{tabular}}
   \footnotesize
  \begin{center}
    \begin{tabular}{|c|c|}
    \hline
      Surface name & Keypoints label\\
    \hline
      Front surface &  \tabincell{c}{0, 1, 2, 3, 4, 5, 6, 8, 49, 51, 52,\\ 53, 54, 55, 56, 57, 58, 59, 60, 61}\\ \hline
      Left surface & \tabincell{c}{7, 9, 10, 11, 12, 13, 14, \\15, 16, 17, 18, 19, 20, 21}\\ \hline
      Rear surface & \tabincell{c}{24, 25, 26, 27, 28, 29, 30, 31, \\32, 33, 34, 35, 62, 63, 64, 65}\\ \hline
      Right surface & \tabincell{c}{36, 37, 38, 39, 40, 41, 42, \\43, 44, 45, 46, 47, 48, 50}\\ \hline
    \end{tabular}
  \caption{\upshape We divided a car into four visible surfaces, and manually define the correspondence between keypoints and surfaces.}
  \label{tab:correspondence}
  \end{center}
 \end{table}

Otherwise, we additionally use $N(c, \ve{M}, \kappa)$, which is a $\kappa$ nearest neighbor function,  to find spatial close car instances and regularize the solved poses.
Specifically, the metric for retrieve neighborhood is the distance between mean coordinates of labelled keypoints. Here we set $\kappa = 2$.

As illustrated in \figref{fig:label_pipeline}, to minimize~\equref{eqn:pose_energy}, we first solve for those cars with dense keypoint annotations, by exhausting all car types. We require that the average re-projection error must be below 5 pixels and the re-projected boundary offset to be within 5 pixels. If more than one cars meet the constraints, we choose the one with minimum re-projection error. We then solve for the cars with fewer keypoint annotations, by using its context information provided by its neighboring cars. After most cars are aligned, we ask human annotators to visually verify and adjust the result before committing to the database.

%% file: approach.tex
\section{Two Baseline Algorithms}
\label{sec:approach} 
Based on ApolloCar3D, we aim to develop strong baseline algorithms to facilitate benchmarking and future research. We first review the most recent literature and then implement two possibly strongest baseline algorithms. 

\paragraph{Existing work on 3D instance recovery from images.} 3D objects are usually recovered from multiple frames, 3D range sensors~\cite{kehl2016deep}, or learning-based methods~\cite{Yu_2018_CVPR,Reddy_2018_CVPR}. 
Nevertheless, addressing 3D instance understanding from a single image in an uncontrolled environment is ill-posed and challenging, thus attracting growing attention. 
With the development of deep CNNs, researchers are able to achieve impressive results with supervised~\cite{fidler20123d,zia2013detailed,miao2016robust,mousavian20173d,tulsiani2015viewpoints,su2015render,xiang2015data,zia2015towards,chen2016monocular,li2016deep,poirson2016fast,massa2016crafting, chabot2017deep, Yang_2018_ECCV} or weakly supervised strategies~\cite{kundu20183d,Pavlakos_2018_CVPR,kanazawa2018learning}. 
Existing works consider to represent an object as a parameterized 3D bounding box~\cite{fidler20123d,su2015render,tulsiani2015viewpoints,poirson2016fast}, coarse wire-frame skeletons~\cite{dingvehicle,li2016deep,wu2016single,zia2013detailed,zeeshan2014cars}, voxels~\cite{choy20163d}, one-hot selection from a small set of exemplar models~\cite{chabot2017deep,mottaghi2015coarse,aubry2014seeing}, and point clouds~\cite{fan2017point}. Category-specific deformable model has also been used for shapes of simple geometry~\cite{kar2015category,kanazawa2018learning}. 

\begin{figure*}[t]
\centering
\includegraphics[width = 0.9\linewidth]{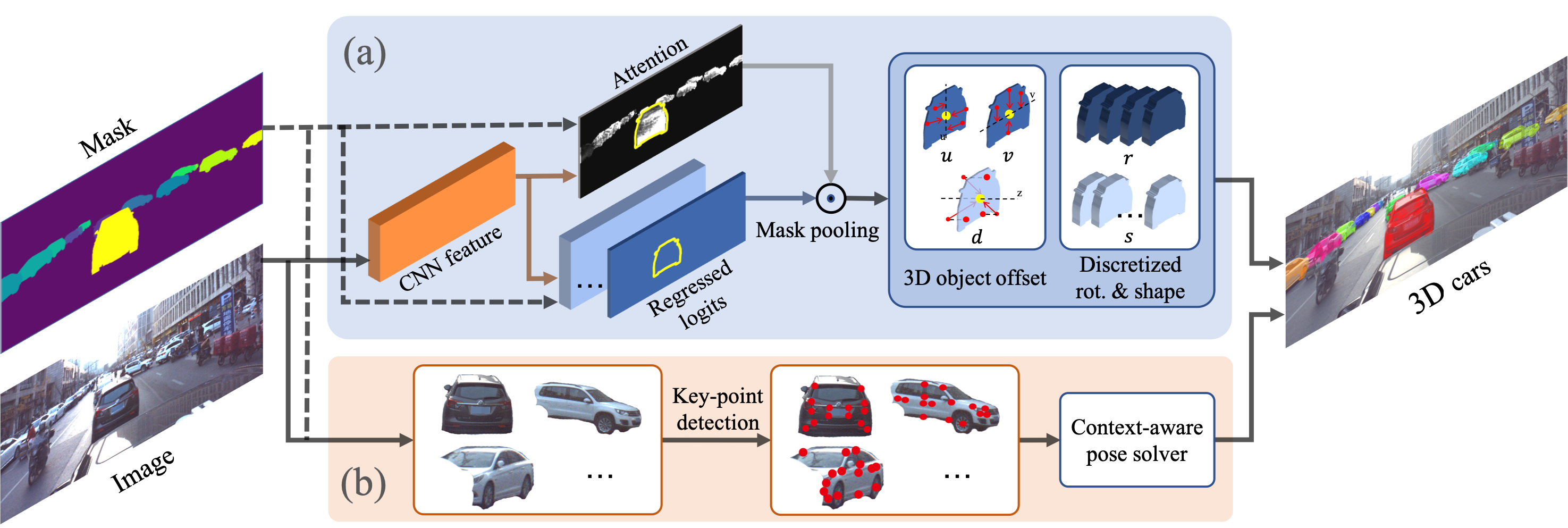}
\caption{Training pipeline for 3D car understanding. Upper (a): direct approach. Bottom (b): key point based approach.}
\label{fig:pipeline}
\end{figure*}

For handling cases of multiple instance, 3D-RCNN~\cite{kundu20183d} and DeepMANTA~\cite{chabot2017deep} are possibly the state-of-the-art techniques by combining 3D shape model with Faster R-CNN~\cite{ren2015faster} detection.
However, due to the lack of high quality dataset, these methods have to rely on 2D masks or wire-frames that are coarse information for supervision. 
Back on ApolloCar3D, in this paper, we adapt their algorithms and conduct supervised training to obtain strong results for benchmarks. Specifically, 3D-RCNN does not consider the car keypoints, which we referred to as direct approach, while DeepMANTA considers keypoints for training and inference, which we call keypoint-based approach. Nevertheless, both algorithms are not open-sourced yet. Therefore, we have to develop our in-house implementation of their methods, serving as baselines in this paper. In addition, we also propose new ideas to improve the baselines, as illustrated in \figref{fig:pipeline}, which we will elaborate later.


Specifically, similar to 3D-RCNN~\cite{kundu20183d}, we assume predicted 2D car masks are given, \eg\ learned through Mask-RCNN~\cite{he2017mask}, and we primarily focus on 3D shape and pose recovery.

\subsection{A Direct Approach}
\label{subsec:direct}
When only car pose and shape are provided, following direct supervision strategy as mentioned in 3D-RCNN~\cite{kundu20183d}, we crop out corresponding features for every car instance from a fully convolutional feature extractor with RoI pooling, and build independent fully connected layers to regress towards its 2D amodal center, allocentric rotation, and PCA-based shape parameters. 
Following the same strategy, the regression output spaces of rotation and shape are discretized.
Nevertheless, for estimating depth, instead of using amodal box and enumerating depth such that the projected mask best fits the box as mentioned in~\cite{kundu20183d}, we use ground truth depths as supervision. Therefore, for our implementation, we replace amodal box regression to depth regression using similar depth discretizing policy as proposed in~\cite{fu2018deep}, which provides state-of-the-art depth estimation from a single image. 

Targeting at detailed shape understanding, we further make two improvements over the original pipeline, as shown in \figref{fig:pipeline}(a).
First, as mentioned in~\cite{kundu20183d}, estimating object 3D shape and pose are distortion-sensitive, and RoI pooling is equivalent to making perspective distortion of an instance in the image, which negatively impact the estimation. 3D-RCNN~\cite{kundu20183d} induces infinity homography to handle the problem. 
In our case, we replace RoI pooling to a fully convolutional architecture, and perform per-pixel regression towards our pose and shape targets, which is simpler yet more effective. Then we aggregate all the predictions inside the given instance mask with a ``self-attention'' policy as commonly used for feature selection~\cite{vaswani2017attention}. Formally, let $\ve{X} \in \mathbb{R}^{h\times w \times c}$ be the feature map, and the output for car instance $i$ is computed as, 
\begin{align}
    \ve{o}_i = \sum_{\ve{x}}\ve{M}_{\ve{x}}^i (\kappa_o * \ve{X} + \ve{b}_o)_\ve{x} \ve{A}_\ve{x}
    \label{eqn:mask_pooling}
\end{align}
where $\ve{o}_i$ is the logits of discretized 3D representation, $\ve{x}$ is a pixel in the image, $\ve{M}^i$ is a binary mask of object $i$, $\kappa_o \in \mathbb{R}^{k_l \times k \times c \times b}$ is the kernels used for predicting outputs, and $\ve{A} \in \mathbb{R}^{h\times w\times 1}$ is the attention map. $b$ is the number of bins for discretization following~\cite{kundu20183d}. We call feature aggregation as mask pooling since it selects the most important information within each object mask.

Secondly, as shown in our pipeline, for estimating car translation, \ie~its amodal center $\ve{c}_a = [c_x, c_y]$ and depth $d_c$, instead of using the same target for every pixel in a car mask, we propose to output a 3D offset at each pixel \wrt the 3D car center, which provides stronger supervision and helps learn more robust networks. 
Previously, inducing relative position of object instances has also been shown to be effective in instance segmentation~\cite{uhrig2018box2pix,li2016fully}.
Formally, let $\ve{c} = [d_c(c_x-u_x)/f_x,d_c(c_y-u_y)/f_y,d_c]$ be the 3D car center, and our 3D offset for a pixel $\ve{x} = [x, y]$ is defined as $\ve{f}^3 = \ve{x}^3 - \ve{c}$, where $\ve{x}^3 = [d(x-u_x)/f_x,d(y-u_y)/f_y,d]$, and $d$ is the estimated depth at $\ve{x}$. 
In principle, 3D offset estimation is equivalent to jointly computing per-pixel 2D offset respect to the amodal center, \ie~ $\ve{x} - \ve{c}_a = [u, v]^T$ and a relative depth to the center depth, \ie~$d - d_c$. We adopt such a factorized representation for model center estimation, and 
the 3D model center can then be recovered by 
\begin{align}
 \ve{c}_a = \sum_\ve{x} \ve{A}_\ve{x} (\ve{x} + \ve{f}^3_{x,y}), d_c = \sum_\ve{x} \ve{A}_\ve{x} (d_\ve{x} + \ve{f}^3_d)
\label{eqn:center}
\end{align}
where $ve{A}_\ve{x}$ is the attention at $\ve{x}$, which is used for output aggregation in \equref{eqn:mask_pooling}.
In our experiments in \secref{sec:exp}, we show that the two strategies provide improvements over the original baseline results.

%% file: kp_based.tex
\subsection{A Keypoint-based Approach}
When sufficient 2D keypoints from each car are available (\eg as in \figref{fig:pipeline}(b)), we develop a simple baseline algorithm, inspired by DeepMANTA~\cite{chabot2017deep}, to align 3D car pose via 2D-3D matching. 

Different from ~\cite{chabot2017deep}, our 3D car models have much more geometric details and come with the absolute scale, and our 2d keypoints have more precise annotations. Here, we adopt the CPM ~\cite{wei2016convolutional} -- a state-of-the-art 2d keypoint detector despite the algorithm was originally developed for human pose estimation. We extend it to 2d car keypoint detection and find it works well. 

One advantage of using 2d keypoint prediction over our baseline-1 \ie the ``direct approach'' in \secref{subsec:direct}, is that,  we do not have to regress the global depth or scale -- the estimation of which by networks is in general not very reliable.  Instead of feeding the full image into the network, we crop out each car region in the image for 2d keypoint detection. This is especially useful for images in ApolloScape~\cite{huang2018apolloscape}, which have a large number of cars of small size.

Borrowing the context-aware constraints from our annotation process, once we have enough detected keypoints, we first solve the easy cases where a car is less occluded using EPnP\cite{lepetit2009epnp}, then we propagate the information to neighboring cars until all car pose and shapes are found to be consistent with each other \wrt the co-planar constraints via optimizing \equref{eqn:pose_energy}. We referred our car pose solver with co-planar constraints as \textbf{context-aware solver}.

%% file: exp_arxiv.tex
\section{Experiments}
\label{sec:exp}
This section provides key implementation details, our newly proposed evaluation metric, and experiment results. In total, we have experimented on 5,277 images, split to 4,036 for training, 200 for validation, and 1,041 for testing. We sample images for each set following the distribution illustrated in~\figref{fig:stat}. The goal is to make sure that the testing data cover a wide range of both easy and difficult scenarios.
\paragraph{Implementation details.} 
Due to the lacking of publicly available source codes, we re-implemented 3D-RCNN~\cite{kundu20183d} for 3D car understanding without using keypoints, and DeepMANTA~\cite{chabot2017deep} which requires key points annotation.
For training Mask-RCNN, we downloaded the code from GitHub implemented by an autonomous driving company~\footnote{https://github.com/TuSimple/mx-maskrcnn}.
We adopted the fully convolutional features from DeepLabv3~\cite{chen2018deeplab} with Xception65~\cite{chollet2017xception} network and follow the same training policy. For DeepMANTA, we used the key point prediction methods from CPM~\cite{chencascaded}. With 4,036 training images, we obtained about 40,000 labeled vehicles with 2D keypoints, used to train a CPM~\cite{chencascaded} (with 5 stages of CPM, and VGG-16 initialization).

\begin{table}[t]
   \footnotesize
  \begin{center}
    \begin{tabular}{|c|c|c|}
    \hline
      Method & mean pixel error& detection rate\\
    \hline
      CPM~\cite{wei2016convolutional} & $4.39 (px)$ & $75.41\%$\\ \hline
      Human label & $2.67 (px)$ & $92.40 \%$ \\ \hline
    \end{tabular}
  \caption{\upshape Keypoints accuracy.}
  \label{tab:kp_accuracy}
  \end{center}
 \end{table}

\begin{table*}
\center
\setlength\tabcolsep{7pt}
\fontsize{8.5}{9}\selectfont
\begin{tabular}{lcc|ccc|ccc|c}
\toprule[0.13em]
\setlength{\tabcolsep}{10pt}\\
\multicolumn{1}{c}{\multirow{2}{*}{Methods}} &
\multicolumn{1}{c}{\multirow{2}{*}{Mask}} &
\multicolumn{1}{c}{\multirow{2}{*}{wKP}} &
\multicolumn{3}{|c|}{A3DP-Abs} &
\multicolumn{3}{c|}{A3DP-Rel} &
\multicolumn{1}{c}{\multirow{2}{*}{Time(s)}}\\
\cmidrule{4-9}
\multicolumn{3}{c|}{} & mean & c-l & c-s  & mean & c-l   & c-s &\multicolumn{1}{c}{} \\
\hline
\multicolumn{1}{l|}{3D-RCNN$^*$~\cite{kundu20183d}} & gt &- & $16.44$  & $29.70$ & $19.80$ &$10.79$ & $17.82$ & $11.88$ & 0.29s\\
\multicolumn{1}{l|}{ + MP}    & gt &- & $16.73$ & $29.70$ & $18.81$ & $10.10$ & $18.81$ & $11.88$ & 0.32s \\
\multicolumn{1}{l|}{ + MP + OF}    &gt &- & $17.52$  & $30.69$ & $20.79$ &$13.66$ & $19.80$ & $13.86$ & 0.34s\\
\multicolumn{1}{l|}{ + MP + OF}    & pred. &- & $15.15$  & $28.71$ & $17.82$ & $11.49$ & $17.82$ & $11.88$ & 0.34s\\
\hline
\multicolumn{1}{l|}{DeepMANTA$^*$~\cite{chabot2017deep}}  &gt & \checkmark & $20.10$ & $30.69$ & $23.76$ & $16.04$ & $23.76$& $19.80$ & 3.38s\\
\multicolumn{1}{l|}{+ CA-solver}  &gt &\checkmark  & $\mathbf{21.57}$ & $\mathbf{32.62}$ & $\mathbf{26.73}$ & $\mathbf{17.52}$ & $\mathbf{26.73}$ & $\mathbf{20.79}$ & 7.41s\\
\multicolumn{1}{l|}{+ CA-solver} & pred. &\checkmark & $20.39$ & $31.68$ & $24.75$ & $16.53$ & $24.75$ & $19.80$ & 8.5s\\

\hline
\multicolumn{1}{l|}{Human} & gt &\checkmark & $38.22$ & $56.44$ & $49.50$ & $33.27$ & $51.49$ & $41.58$ & 607.41s\\
\toprule[0.13 em]
\end{tabular}
\caption{\small Comparison among baseline algorithms.  $*$ means in-house implementation. ``Mask'' means the provided mask for 3D understanding (``gt'' means ground truth mask and ``pred.'' means Mask-RCNN mask). ``wKP'' means using keypoint predictions. ``c-l'' indicates results from loose criterion, and ``c-s'' indicates results from strict criterion. ``MP'' stands for mask pooling and ``OF'' stands for offset flow. ``CA-solver'' stands for context-aware 3D pose solver. ``Times(s)'' indicates the average inference times cost for processing each image. }
\label{tab:results}
\end{table*}
\paragraph{Evaluation metrics.} Similar to the detection task, the average precision (AP)~\cite{pascal-voc-2012} is usually used for evaluating 3D object understanding. However, the similarity is measured using 3D bounding box IoU~\cite{geiger2012we} with orientation (average orientation similarity (AOS)~\cite{geiger2012we}) or 2D bounding box with viewpoint (average viewpoint precision (AVP)~\cite{xiang2014beyond}).
Unfortunately, those metrics can only measure very coarse 3D properties, yet object shape has not been considered jointly with 3D rotation and translation.

Mesh distance~\cite{Stutz2018CVPR} and voxel IoU~\cite{di20173d} are usually used to evaluate 3D shape reconstruction. In our case, a car model is mostly compact, thus we consider comparing projection masks of two models following the idea of visual hull representation~\cite{matusik2000image}. Specifically, we sample 100 orientations at yaw angular direction and project each view of the model to an image with a resolution of \by{1280}{1280}. We use the mean IoU over all views as the car shape similarity metric. For evaluating rotation and translation, we follow the metrics commonly used for camera pose estimation~\cite{geiger2012we}. In summary, the criteria for judging a true positive given a set of thresholds is defined as
\begin{align}
    c_{shape} &= \frac{1}{|V|}\sum\nolimits_{v\in \hua{V}}IoU(\ve{P}(s_i), \ve{P}(s_i^*))_v \geq \delta_s, \nonumber \\
    c_{trans} &= |\ve{t}_i - \ve{t}^*_i|_2 \leq \delta_t, \nonumber \\
    c_{rot} &= \arccos(|\ve{q}(\ve{r}_i) \cdot \ve{q}(\ve{r}^*_i)|) \leq \delta_r,
\end{align}
where $s, \ve{t}, \ve{r}$ are the shape ID, translation, and rotation of a predicted 3D car instance.

In addition, a single set of true positive thresholds used by AOS or AVP, \eg\ IoU $\geq 0.5$, and rotation $\leq \pi/6$, is not sufficient to evaluate detected results thoroughly~\cite{geiger2012we}. Here, following the metric of MS COCO~\cite{lin2014microsoft},  we propose to use multiple sets of thresholds from loose to strict for evaluation. Specifically, the thresholds used in our results for all levels of ifficulty are $\{\delta_s\} = [0.5:0.05:0.95], \{\delta_t\} = [2.8:0.3:0.1], \{\delta_r\} = [\pi/6:\pi/60:\pi/60]$, where $[a:i:b]$ indicates a set of discrete thresholds sampled in a line space from $a$ to $b$ with an interval of $i$. Similar to MSCOCO, we select one loose criterion $\ve{c}-l = [0.5, 2.8, \pi/6]$ and one strict criterion $\ve{c}-l = [0.75, 1.4, \pi/12]$ to diagnose the performance of different algorithms. Note that in our metrics, we only evaluate instances with depth less than $100m$ as we would like to focus on cars that are more immediately relevant to our autonomous driving task.

Finally, in self-driving scenarios that are safety critical, we commonly care nearby cars rather than those far away. Therefore, we further propose to use a relative error metric for evaluating translation following the ``AbsRel'' commonly used in depth evaluation~\cite{geiger2012we}. Formally, we change the criteria of $c_{trans}$ to $|\ve{t}_i - \ve{t}^*_i|/\ve{t}^*_i \leq \delta^*_t$, and set the thresholds to $\{\delta^*_t\} = [0.10:0.01:0.01]$.
We call our evaluation metric with absolute translation thresholds as ``A3DP-Abs'', and the one with relative translation thresholds as ``A3DP-Rel'', and we report the results under both metrics in our later experiments.

\begin{figure*}[t]
\begin{center}
\includegraphics[width=0.9\textwidth]{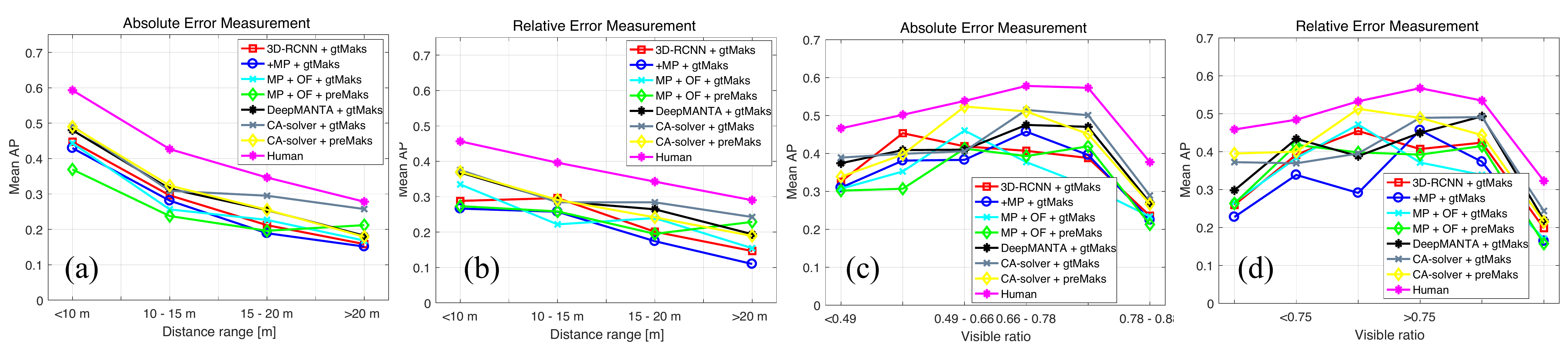}
\end{center}
\caption{3D understanding results of various algorithms \wrt different factors causing false estimation. (a) A3DP-Abs v.s distance, (b) A3DP-Rel v.s distance, (c) A3DP-Abs v.s occlusion, (d) A3DP-Abs v.s occlusion. }
\label{fig:analysis}
\end{figure*}

\begin{figure*}
\centering
\includegraphics[width=0.9\linewidth]{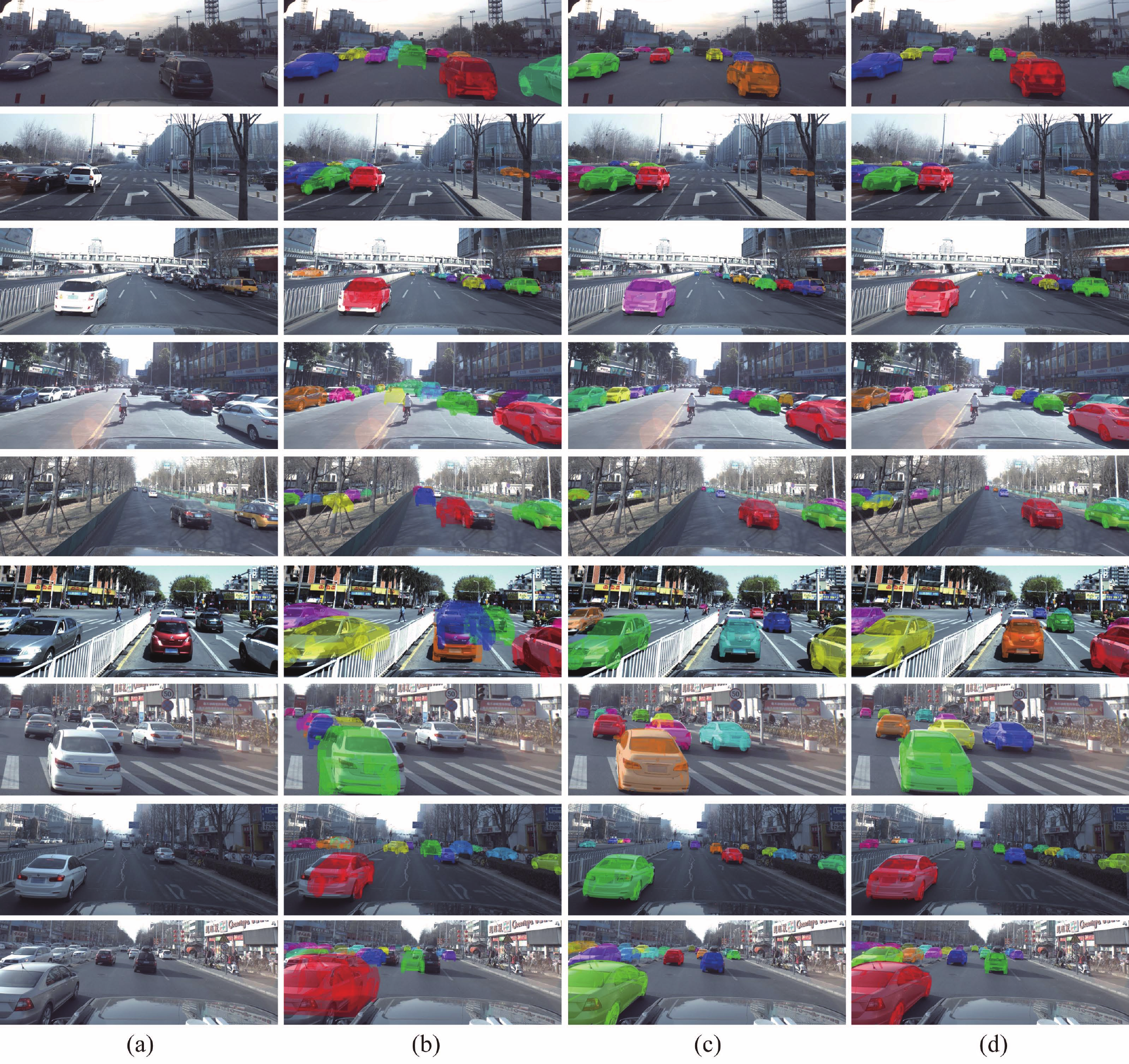}
\caption{Visualization results of different approaches, in which (a) the input image, (b) and (c) are the results with direct regression method and key points-based method with context constraint. (d) gives the ground truth results.}
\label{fig:ablation}
\end{figure*}


\subsection{Quantitative Results}
In this section, we compare against our baseline algorithms with the method presented in \secref{sec:approach} by progressively adding our proposed components and losses. \tabref{tab:results} shows the comparison results. For direct regression approach, our baseline algorithm ``3D-RCNN'' provides regression towards translation, allocentric rotation, and car shape parameters. We further extend the baseline method by adding mask pooling (MP) and offset flow (OF). We observe from the table that, swapping RoI pooling for mask pooling moderately improves the results while offset flow brings significant boost. They together help avoiding geometric distortions from regular RoI pooling and bring attention mechanism to focus on relevant regions.

For the keypoint-based method, ``DeepMANTA'' shows the results by using our detected key points and solving with PnP for each car individually, yielding reasonable performance. ``+CA-solver'' means for cars without sufficient detected key points, we employ our context-aware solver for inference, which provides around $1.5\%$ improvement. For both methods, switching ground truth mask to segmentation from Mask R-CNN gives little drop of the performance, demonstrating the high quality of Mask R-CNN results.

Finally, we train a new group of labellers, and ask them to re-label the keypoints on our validation set, which are passed through our context-aware 3D solver. We denote these results as ``human'' performance. We can see there is a clear gap ($\sim 10\%$) between algorithms with human. However, even the accuracy for humans is still not satisfying. After checking the results, we found that this is primarily because humans cannot accurately memorize the semantic meaning of all the 66 keypoints, yielding wrongly solved poses. We conjecture this could be fixed by rechecking and refinement, possibly leading to improved performance.

Tab.~\ref{tab:kp_accuracy} shows the accuracy of 2d keypoints. For each predicted keypoint, if its distance to ground truth keypoint is less than $10(pixel)$, we regard it as positive, otherwise, it is regarded as negative. We first crop out each car using its ground truth mask, then use CPM~\cite{wei2016convolutional} to train the 2d keypoints detector. 
The detection rate is $75.41$ $\%$(rate of number of positive keypoints and all ground truth), and the mean pixel error is $4.39$ $px$.
We also show the accuracy of human labeled keypoints. The detection rate of human labeled 2d keypoints is $92.40\%$, and the mean pixel error of detected 2d keypoints is $2.67(pixel)$.
As discussed in the paper, the mis-labelling of human is primarily because humans cannot accurately memorize the semantic meaning of all the 66 keypoints. However, it is still much better than a trained CPM keypoint detector because the robustness of human with respect to appearance and occlusion changes.

\subsection{Qualitative Results}
Some qualitative results are visualized in \figref{fig:ablation}. From the two examples, we can find that the additional key point predictions provide more accurate 3D estimation than direct method due to the use of geometric constraints and inter-car relationship constraints. In particular, for the direct method, most errors occur in depth prediction. It can be explained by the nature of the method that the method predicts the global 3D property of depth purely based on object appearance in 2D, which is ill-posed and error-prone. However, thanks to the use of reliable masks, the method discovers more cars than the keypoint-based counterpart. For the keypoint-based approach, we are able to show that correctly detected keypoints are extremely successful at constraining car poses, while failed or missing keypoint estimation, especially for cars of unusual appearance, will lead to missing detection of cars or wrong solution for poses.

\vspace{-0.05in}
\subsection{Result Analysis}
To analyze the performance of different approaches, we evaluate them separately on various distances and occlusion ratios. Detailed results are shown in \figref{fig:analysis}. Checking \figref{fig:analysis}(a, b), as expected, we can find that the estimation accuracy decreases with farther distances, and the gap between human and algorithm narrows in the distance.
In addition, after checking \figref{fig:analysis}(c, d) for occlusion, we discover that the performance also drops with increasing the occlusion ratio. However, we observe that the performance on non-occluded cars is the worst on average among all occlusion patterns. This is because most cars which experience little occlusion are from large distance and of small scale, while cars close-by are more often occluded. 

%% file: 3D_car_arxiv.bbl
\begin{thebibliography}{10}\itemsep=-1pt

\bibitem{aubry2014seeing}
M.~Aubry, D.~Maturana, A.~A. Efros, B.~C. Russell, and J.~Sivic.
\newblock Seeing 3d chairs: exemplar part-based 2d-3d alignment using a large
  dataset of cad models.
\newblock In {\em Proceedings of the IEEE conference on computer vision and
  pattern recognition}, pages 3762--3769, 2014.

\bibitem{canal2016real}
G.~Canal, S.~Escalera, and C.~Angulo.
\newblock A real-time human-robot interaction system based on gestures for
  assistive scenarios.
\newblock {\em Computer Vision and Image Understanding}, 149:65--77, 2016.

\bibitem{chabot2017deep}
F.~Chabot, M.~Chaouch, J.~Rabarisoa, C.~Teuli{\`e}re, and T.~Chateau.
\newblock Deep manta: A coarse-to-fine many-task network for joint 2d and 3d
  vehicle analysis from monocular image.
\newblock In {\em Proc. IEEE Conf. Comp. Vis. Patt. Recogn.}, pages 2040--2049,
  2017.

\bibitem{chang2015shapenet}
A.~X. Chang, T.~Funkhouser, L.~Guibas, P.~Hanrahan, Q.~Huang, Z.~Li,
  S.~Savarese, M.~Savva, S.~Song, H.~Su, et~al.
\newblock Shapenet: An information-rich 3d model repository.
\newblock {\em arXiv preprint arXiv:1512.03012}, 2015.

\bibitem{chen2018deeplab}
L.-C. Chen, G.~Papandreou, I.~Kokkinos, K.~Murphy, and A.~L. Yuille.
\newblock Deeplab: Semantic image segmentation with deep convolutional nets,
  atrous convolution, and fully connected crfs.
\newblock {\em IEEE transactions on pattern analysis and machine intelligence},
  40(4):834--848, 2018.

\bibitem{chen2016monocular}
X.~Chen, K.~Kundu, Z.~Zhang, H.~Ma, S.~Fidler, and R.~Urtasun.
\newblock Monocular 3d object detection for autonomous driving.
\newblock In {\em Proceedings of the IEEE Conference on Computer Vision and
  Pattern Recognition}, pages 2147--2156, 2016.

\bibitem{chencascaded}
Y.~Chen, Z.~Wang, Y.~Peng, Z.~Zhang, and G.~Y.~J. Sun.
\newblock Cascaded pyramid network for multi-person pose estimation.

\bibitem{chollet2017xception}
F.~Chollet.
\newblock Xception: Deep learning with depthwise separable convolutions.
\newblock {\em arXiv preprint}, pages 1610--02357, 2017.

\bibitem{choy20163d}
C.~B. Choy, D.~Xu, J.~Gwak, K.~Chen, and S.~Savarese.
\newblock 3d-r2n2: A unified approach for single and multi-view 3d object
  reconstruction.
\newblock In {\em Proc. Eur. Conf. Comp. Vis.}, 2016.

\bibitem{dai2017scannet}
A.~Dai, A.~X. Chang, M.~Savva, M.~Halber, T.~A. Funkhouser, and M.~Nie{\ss}ner.
\newblock Scannet: Richly-annotated 3d reconstructions of indoor scenes.
\newblock In {\em Proc. IEEE Conf. Comp. Vis. Patt. Recogn.}

\bibitem{deng2009imagenet}
J.~Deng, W.~Dong, R.~Socher, L.-J. Li, K.~Li, and L.~Fei-Fei.
\newblock Imagenet: A large-scale hierarchical image database.
\newblock In {\em Proc. IEEE Conf. Comp. Vis. Patt. Recogn.}, pages 248--255.
  Ieee, 2009.

\bibitem{di20173d}
X.~Di and P.~Yu.
\newblock 3d reconstruction of simple objects from a single view silhouette
  image.
\newblock {\em arXiv preprint arXiv:1701.04752}, 2017.

\bibitem{Reddy_2018_CVPR}
N.~Dinesh~Reddy, M.~Vo, and S.~G. Narasimhan.
\newblock Carfusion: Combining point tracking and part detection for dynamic 3d
  reconstruction of vehicles.
\newblock In {\em Proc. IEEE Conf. Comp. Vis. Patt. Recogn.}, June 2018.

\bibitem{dingvehicle}
W.~Ding, S.~Li, G.~Zhang, X.~Lei, H.~Qian, and Y.~Xu.
\newblock Vehicle pose and shape estimation through multiple monocular vision.

\bibitem{engelmann2016joint}
F.~Engelmann, J.~St{\"u}ckler, and B.~Leibe.
\newblock Joint object pose estimation and shape reconstruction in urban street
  scenes using 3d shape priors.
\newblock In {\em German Conference on Pattern Recognition}, pages 219--230.
  Springer, 2016.

\bibitem{pascal-voc-2012}
M.~Everingham, L.~Van~Gool, C.~K.~I. Williams, J.~Winn, and A.~Zisserman.
\newblock The {PASCAL} {V}isual {O}bject {C}lasses {C}hallenge 2012 {(VOC2012)}
  {R}esults.
\newblock
  http://www.pascal-network.org/challenges/VOC/voc2012/workshop/index.html.

\bibitem{fan2017point}
H.~Fan, H.~Su, and L.~J. Guibas.
\newblock A point set generation network for 3d object reconstruction from a
  single image.
\newblock In {\em Proc. IEEE Conf. Comp. Vis. Patt. Recogn.}

\bibitem{fidler20123d}
S.~Fidler, S.~Dickinson, and R.~Urtasun.
\newblock 3d object detection and viewpoint estimation with a deformable 3d
  cuboid model.
\newblock In {\em Proc. Adv. Neural Inf. Process. Syst.}, pages 611--619, 2012.

\bibitem{fischler1981random}
M.~A. Fischler and R.~C. Bolles.
\newblock Random sample consensus: a paradigm for model fitting with
  applications to image analysis and automated cartography.
\newblock {\em Communications of the ACM}, 24(6):381--395, 1981.

\bibitem{fu2018deep}
H.~Fu, M.~Gong, C.~Wang, K.~Batmanghelich, and D.~Tao.
\newblock Deep ordinal regression network for monocular depth estimation.
\newblock In {\em Proceedings of the IEEE Conference on Computer Vision and
  Pattern Recognition}, pages 2002--2011, 2018.

\bibitem{geiger2012we}
A.~Geiger, P.~Lenz, and R.~Urtasun.
\newblock Are we ready for autonomous driving? the kitti vision benchmark
  suite.
\newblock In {\em Proc. IEEE Conf. Comp. Vis. Patt. Recogn.}, pages 3354--3361.
  IEEE, 2012.

\bibitem{he2017mask}
K.~He, G.~Gkioxari, P.~Doll{\'a}r, and R.~Girshick.
\newblock Mask r-cnn.
\newblock In {\em Proc. IEEE Int. Conf. Comp. Vis.}, pages 2980--2988. IEEE,
  2017.

\bibitem{huang2018apolloscape}
X.~Huang, X.~Cheng, Q.~Geng, B.~Cao, D.~Zhou, P.~Wang, Y.~Lin, and R.~Yang.
\newblock The apolloscape dataset for autonomous driving.
\newblock In {\em Proceedings of the IEEE Conference on Computer Vision and
  Pattern Recognition Workshops}, pages 954--960, 2018.

\bibitem{kanazawa2018learning}
A.~Kanazawa, S.~Tulsiani, A.~A. Efros, and J.~Malik.
\newblock Learning category-specific mesh reconstruction from image
  collections.
\newblock In {\em Proc. Eur. Conf. Comp. Vis.}, 2018.

\bibitem{kar2015category}
A.~Kar, S.~Tulsiani, J.~Carreira, and J.~Malik.
\newblock Category-specific object reconstruction from a single image.
\newblock In {\em Proc. IEEE Conf. Comp. Vis. Patt. Recogn.}, pages 1966--1974,
  2015.

\bibitem{kehl2016deep}
W.~Kehl, F.~Milletari, F.~Tombari, S.~Ilic, and N.~Navab.
\newblock Deep learning of local rgb-d patches for 3d object detection and 6d
  pose estimation.
\newblock In {\em Proc. Eur. Conf. Comp. Vis.}, pages 205--220. Springer, 2016.

\bibitem{kulkarni2015deep}
T.~D. Kulkarni, W.~F. Whitney, P.~Kohli, and J.~Tenenbaum.
\newblock Deep convolutional inverse graphics network.
\newblock In {\em Proc. Adv. Neural Inf. Process. Syst.}, pages 2539--2547,
  2015.

\bibitem{kundu20183d}
A.~Kundu, Y.~Li, and J.~M. Rehg.
\newblock 3d-rcnn: Instance-level 3d object reconstruction via
  render-and-compare.
\newblock In {\em Proc. IEEE Conf. Comp. Vis. Patt. Recogn.}, pages 3559--3568,
  2018.

\bibitem{leibe2003analyzing}
B.~Leibe and B.~Schiele.
\newblock Analyzing appearance and contour based methods for object
  categorization.
\newblock In {\em Proc. IEEE Conf. Comp. Vis. Patt. Recogn.}, volume~2, pages
  II--409, 2003.

\bibitem{leonard2012directed}
J.~J. Leonard and H.~F. Durrant-Whyte.
\newblock {\em Directed sonar sensing for mobile robot navigation}, volume 175.
\newblock Springer Science \& Business Media, 2012.

\bibitem{lepetit2009epnp}
V.~Lepetit, F.~Moreno-Noguer, and P.~Fua.
\newblock Epnp: An accurate o (n) solution to the pnp problem.
\newblock {\em Int. J. Comp. Vis.}, 81(2):155, 2009.

\bibitem{li2016deep}
C.~Li, M.~Z. Zia, Q.-H. Tran, X.~Yu, G.~D. Hager, and M.~Chandraker.
\newblock Deep supervision with shape concepts for occlusion-aware 3d object
  parsing.
\newblock {\em arXiv preprint arXiv:1612.02699}, 2016.

\bibitem{li2016fully}
Y.~Li, H.~Qi, J.~Dai, X.~Ji, and Y.~Wei.
\newblock Fully convolutional instance-aware semantic segmentation.
\newblock {\em arXiv preprint arXiv:1611.07709}, 2016.

\bibitem{liang2018deep}
M.~Liang, B.~Yang, S.~Wang, and R.~Urtasun.
\newblock Deep continuous fusion for multi-sensor 3d object detection.
\newblock In {\em Proc. Eur. Conf. Comp. Vis.}, pages 641--656, 2018.

\bibitem{lim2013parsing}
J.~J. Lim, H.~Pirsiavash, and A.~Torralba.
\newblock Parsing ikea objects: Fine pose estimation.
\newblock In {\em Proc. IEEE Int. Conf. Comp. Vis.}, pages 2992--2999, 2013.

\bibitem{lin2014microsoft}
T.-Y. Lin, M.~Maire, S.~Belongie, J.~Hays, P.~Perona, D.~Ramanan,
  P.~Doll{\'a}r, and C.~L. Zitnick.
\newblock Microsoft coco: Common objects in context.
\newblock In {\em Proc. Eur. Conf. Comp. Vis.}, pages 740--755. Springer, 2014.

\bibitem{lopez2010icaro}
R.~Lopez-Sastre, C.~Redondo-Cabrera, P.~Gil-Jimenez, and S.~Maldonado-Bascon.
\newblock Icaro: image collection of annotated real-world objects, 2010.

\bibitem{massa2016crafting}
F.~Massa, R.~Marlet, and M.~Aubry.
\newblock Crafting a multi-task cnn for viewpoint estimation.
\newblock {\em arXiv preprint arXiv:1609.03894}, 2016.

\bibitem{matusik2000image}
W.~Matusik, C.~Buehler, R.~Raskar, S.~J. Gortler, and L.~McMillan.
\newblock Image-based visual hulls.
\newblock In {\em Proceedings of the 27th annual conference on Computer
  graphics and interactive techniques}, pages 369--374. ACM
  Press/Addison-Wesley Publishing Co., 2000.

\bibitem{mcauley2012image}
J.~McAuley and J.~Leskovec.
\newblock Image labeling on a network: using social-network metadata for image
  classification.
\newblock In {\em Proc. Eur. Conf. Comp. Vis.}, pages 828--841. Springer, 2012.

\bibitem{menze2015object}
M.~Menze and A.~Geiger.
\newblock Object scene flow for autonomous vehicles.
\newblock In {\em Proc. IEEE Conf. Comp. Vis. Patt. Recogn.}, pages 3061--3070,
  2015.

\bibitem{Menze2015ISA}
M.~Menze, C.~Heipke, and A.~Geiger.
\newblock Joint 3d estimation of vehicles and scene flow.
\newblock In {\em ISPRS Workshop on Image Sequence Analysis (ISA)}, 2015.

\bibitem{miao2016robust}
Y.~Miao, X.~Tao, and J.~Lu.
\newblock Robust 3d car shape estimation from landmarks in monocular image.
\newblock In {\em Proc. Brit. Mach. Vis. Conf.}, 2016.

\bibitem{moreels2007evaluation}
P.~Moreels and P.~Perona.
\newblock Evaluation of features detectors and descriptors based on 3d objects.
\newblock {\em Int. J. Comp. Vis.}, 73(3):263--284, 2007.

\bibitem{mottaghi2015coarse}
R.~Mottaghi, Y.~Xiang, and S.~Savarese.
\newblock A coarse-to-fine model for 3d pose estimation and sub-category
  recognition.
\newblock In {\em Proc. IEEE Conf. Comp. Vis. Patt. Recogn.}, pages 418--426,
  2015.

\bibitem{mousavian20173d}
A.~Mousavian, D.~Anguelov, J.~Flynn, and J.~Ko{\v{s}}eck{\'a}.
\newblock 3d bounding box estimation using deep learning and geometry.
\newblock In {\em Proc. IEEE Conf. Comp. Vis. Patt. Recogn.}, pages 5632--5640.
  IEEE, 2017.

\bibitem{ozuysal2009pose}
M.~Ozuysal, V.~Lepetit, and P.~Fua.
\newblock Pose estimation for category specific multiview object localization.
\newblock In {\em Proc. IEEE Conf. Comp. Vis. Patt. Recogn.}, pages 778--785.
  IEEE, 2009.

\bibitem{Pavlakos_2018_CVPR}
G.~Pavlakos, L.~Zhu, X.~Zhou, and K.~Daniilidis.
\newblock Learning to estimate 3d human pose and shape from a single color
  image.
\newblock In {\em Proc. IEEE Conf. Comp. Vis. Patt. Recogn.}, June 2018.

\bibitem{poirson2016fast}
P.~Poirson, P.~Ammirato, C.-Y. Fu, W.~Liu, J.~Kosecka, and A.~C. Berg.
\newblock Fast single shot detection and pose estimation.
\newblock In {\em 3dv}, pages 676--684. IEEE, 2016.

\bibitem{ren2015faster}
S.~Ren, K.~He, R.~Girshick, and J.~Sun.
\newblock Faster r-cnn: Towards real-time object detection with region proposal
  networks.
\newblock In {\em Advances in neural information processing systems}, pages
  91--99, 2015.

\bibitem{russell2009building}
B.~C. Russell and A.~Torralba.
\newblock Building a database of 3d scenes from user annotations.
\newblock In {\em Proc. IEEE Conf. Comp. Vis. Patt. Recogn.}, pages 2711--2718.
  IEEE, 2009.

\bibitem{savarese20073d}
S.~Savarese and L.~Fei-Fei.
\newblock 3d generic object categorization, localization and pose estimation.
\newblock In {\em Proc. IEEE Conf. Comp. Vis. Patt. Recogn.}, pages 1--8, Oct
  2007.

\bibitem{Stutz2018CVPR}
D.~Stutz and A.~Geiger.
\newblock Learning 3d shape completion from laser scan data with weak
  supervision.
\newblock In {\em IEEE Conference on Computer Vision and Pattern Recognition
  (CVPR)}. IEEE Computer Society, 2018.

\bibitem{su2015render}
H.~Su, C.~R. Qi, Y.~Li, and L.~J. Guibas.
\newblock Render for cnn: Viewpoint estimation in images using cnns trained
  with rendered 3d model views.
\newblock In {\em Proc. IEEE Int. Conf. Comp. Vis.}, pages 2686--2694, 2015.

\bibitem{thomas2006towards}
A.~Thomas, V.~Ferrar, B.~Leibe, T.~Tuytelaars, B.~Schiel, and L.~V. Gool.
\newblock Towards multi-view object class detection.
\newblock In {\em Proc. IEEE Conf. Comp. Vis. Patt. Recogn.}, volume~2, pages
  1589--1596, June 2006.

\bibitem{tran2015learning}
D.~Tran, L.~Bourdev, R.~Fergus, L.~Torresani, and M.~Paluri.
\newblock Learning spatiotemporal features with 3d convolutional networks.
\newblock In {\em Proc. IEEE Int. Conf. Comp. Vis.}, pages 4489--4497, 2015.

\bibitem{tulsiani2015viewpoints}
S.~Tulsiani and J.~Malik.
\newblock Viewpoints and keypoints.
\newblock In {\em Proc. IEEE Conf. Comp. Vis. Patt. Recogn.}, pages 1510--1519,
  2015.

\bibitem{uhrig2018box2pix}
J.~Uhrig, E.~Rehder, B.~Fr{\"o}hlich, U.~Franke, and T.~Brox.
\newblock Box2pix: Single-shot instance segmentation by assigning pixels to
  object boxes.
\newblock In {\em IEEE Intelligent Vehicles Symposium (IV)}, 2018.

\bibitem{vaswani2017attention}
A.~Vaswani, N.~Shazeer, N.~Parmar, J.~Uszkoreit, L.~Jones, A.~N. Gomez,
  {\L}.~Kaiser, and I.~Polosukhin.
\newblock Attention is all you need.
\newblock In {\em Advances in Neural Information Processing Systems}, pages
  5998--6008, 2017.

\bibitem{wang20183d}
Y.~Wang, X.~Tan, Y.~Yang, X.~Liu, E.~Ding, F.~Zhou, and L.~S. Davis.
\newblock 3d pose estimation for fine-grained object categories.
\newblock {\em arXiv preprint arXiv:1806.04314}, 2018.

\bibitem{wei2016convolutional}
S.-E. Wei, V.~Ramakrishna, T.~Kanade, and Y.~Sheikh.
\newblock Convolutional pose machines.
\newblock In {\em Proceedings of the IEEE Conference on Computer Vision and
  Pattern Recognition}, pages 4724--4732, 2016.

\bibitem{wu2016single}
J.~Wu, T.~Xue, J.~J. Lim, Y.~Tian, J.~B. Tenenbaum, A.~Torralba, and W.~T.
  Freeman.
\newblock Single image 3d interpreter network.
\newblock In {\em ECCV}, pages 365--382. Springer, 2016.

\bibitem{xiang2015data}
Y.~Xiang, W.~Choi, Y.~Lin, and S.~Savarese.
\newblock Data-driven 3d voxel patterns for object category recognition.
\newblock In {\em Proc. IEEE Conf. Comp. Vis. Patt. Recogn.}, pages 1903--1911,
  2015.

\bibitem{xiang2016objectnet3d}
Y.~Xiang, W.~Kim, W.~Chen, J.~Ji, C.~Choy, H.~Su, R.~Mottaghi, L.~Guibas, and
  S.~Savarese.
\newblock Objectnet3d: A large scale database for 3d object recognition.
\newblock In {\em Proc. Eur. Conf. Comp. Vis.}, pages 160--176. Springer, 2016.

\bibitem{xiang2014beyond}
Y.~Xiang, R.~Mottaghi, and S.~Savarese.
\newblock Beyond pascal: A benchmark for 3d object detection in the wild.
\newblock pages 75--82. IEEE, 2014.

\bibitem{Yang_2018_ECCV}
G.~Yang, Y.~Cui, S.~Belongie, and B.~Hariharan.
\newblock Learning single-view 3d reconstruction with limited pose supervision.
\newblock In {\em Proc. Eur. Conf. Comp. Vis.}, September 2018.

\bibitem{Yu_2018_CVPR}
T.~Yu, J.~Meng, and J.~Yuan.
\newblock Multi-view harmonized bilinear network for 3d object recognition.
\newblock In {\em Proc. IEEE Conf. Comp. Vis. Patt. Recogn.}, June 2018.

\bibitem{zeeshan2014cars}
M.~Zeeshan~Zia, M.~Stark, and K.~Schindler.
\newblock Are cars just 3d boxes?-jointly estimating the 3d shape of multiple
  objects.
\newblock In {\em Proceedings of the IEEE Conference on Computer Vision and
  Pattern Recognition}, pages 3678--3685, 2014.

\bibitem{zia2013detailed}
M.~Z. Zia, M.~Stark, B.~Schiele, and K.~Schindler.
\newblock Detailed 3d representations for object recognition and modeling.
\newblock {\em {IEEE} Trans. Pattern Anal. Mach. Intell.}, 35(11):2608--2623,
  2013.

\bibitem{zia2015towards}
M.~Z. Zia, M.~Stark, and K.~Schindler.
\newblock Towards scene understanding with detailed 3d object representations.
\newblock {\em Int. J. Comp. Vis.}, 112(2):188--203, 2015.

\end{thebibliography}
